\documentclass{article}

\PassOptionsToPackage{numbers}{natbib}

 \usepackage[preprint]{neurips_2019}

\usepackage[utf8]{inputenc} 
\usepackage[T1]{fontenc}    
\usepackage{hyperref}       
\usepackage{url}            
\usepackage{booktabs}       
\usepackage{amsfonts}       
\usepackage{nicefrac}       
\usepackage{microtype}      

\usepackage{bm}
\usepackage{amsmath,amsthm,amsfonts,amssymb}
\usepackage{wrapfig}
\usepackage{algorithm}
\usepackage{algorithmicx}
\usepackage{algpseudocode}
\usepackage{subcaption}
\usepackage{verbatim}
\newcommand*{\affaddr}[1]{#1} 
\newcommand*{\affmark}[1][*]{\textsuperscript{#1}}
\newcommand*{\email}[1]{\texttt{#1}}

\usepackage{graphicx}

\newtheorem{theorem}{Theorem}

\title{Convolutional Geometric Matrix Completion}

\author{
	Kai-Lang Yao\affmark[1,2], Wu-Jun Li\affmark[1,2], Jianbo Yang\affmark[3] {\normalfont and} Xinyan Lu\affmark[3]\\
	\affaddr{\affmark[1]National Key Laboratory for Novel Software Technology}\\
	\affaddr{\affmark[2]Deptpartment of Computer Science and Technology,
		Nanjing University, China}\\
	\affaddr{\affmark[3]Tencent}\\
	\email{yaokl@lamda.nju.edu.cn,liwujun@nju.edu.cn,\{jianboyang,xinyanlu\}@tencent.com} \\
}

\begin{document}

\maketitle

\begin{abstract}
Geometric matrix completion~(GMC) has been proposed for recommendation by integrating the relationship~(link) graphs among users/items into matrix completion~(MC). Traditional \mbox{GMC} methods typically adopt graph regularization to impose smoothness priors for \mbox{MC}. Recently, geometric deep learning on graphs~(\mbox{GDLG}) is proposed to solve the GMC problem, showing better performance than existing GMC methods including traditional graph regularization based methods. To the best of our knowledge, there exists only one GDLG method for GMC, which is called \mbox{RMGCNN}. RMGCNN combines graph convolutional network~(GCN) and recurrent neural network~(RNN) together for GMC. In the original work of RMGCNN, RMGCNN demonstrates better performance than pure GCN-based method. In this paper, we propose a new GMC method, called \underline{c}onvolutional \underline{g}eometric \underline{m}atrix \underline{c}ompletion~(CGMC), for recommendation with graphs among users/items. CGMC is a pure GCN-based method with a newly designed graph convolutional network. Experimental results on real datasets show that CGMC can outperform other state-of-the-art methods including RMGCNN in terms of both accuracy and speed.
\end{abstract}

\section{Introduction}
Recommender systems~\cite{netflix} have been widely deployed in lots of applications, such as item recommendation on shopping web site, friend recommendation on social web site and so on. There are three main kinds of methods for recommender systems, content-based filtering~\cite{cf2}, collaborative filtering~\cite{cf1} and hybrid methods~\cite{burke02}. Content-based filtering methods recommend new items that are most similar to users' historical favorite items. Collaborative filtering methods use collective ratings to make new recommendations by similar rating patterns between users or items. Hybrid methods integrate the above two methods.

By regarding rows as users, columns as items and entries as ratings on items by users, the task of recommender systems can be formulated as a matrix completion~(MC) problem~\cite{mc, mc1}. MC has attracted lots of attention in recent years. MC models aim to predict the missing entries in a matrix given a small subset of observed entries. Under the low-rank setting, \cite{mc, mc1} have proved that matrix can be exactly recovered given sufficiently large number of observed entries, although it is a NP-hard problem. One efficient solution for MC problem is to adopt matrix factorization~(MF) techniques~\cite{cabral13,mf,mf1,srmgcnn}.

In many real applications, besides the rating matrix which contains the ratings on items by users, other side information is also available. Typical side information includes attributes of users/items and the relationship~(link) graphs between users/items. Therefore, there have appeared a few works to incorporate the attributes of users/items to boost the performance of matrix completion models \cite{imc1, imc2, imc3, imc4}. Furthermore, geometric matrix completion~(GMC) models~\cite{gmc2,gmc,grals,srmgcnn} have also been proposed for recommendation by integrating the relationship~(link) graphs among users/items into matrix completion. For example, the methods in~\cite{gmc2, gmc1, gmc3,grals} propose to encode the structural~(geometric) information of graphs via graph Laplacian regularization~\cite{laplacian1, laplacian2} which tries to impose smoothness priors on latent factors~(embeddings) of users/items. These graph regularization based methods have shown promising performance in real applications.

Recently, geometric deep learning techniques~\cite{gdnn1,gdnn5,gcnn,gnn1,gnn2,graphsage,gat,gdl,stochastic-gcn,adaptive-gcn,Klicpera19,Haija19} are proposed to learn meaningful representations for geometric structure data, such as graphs and manifolds. In particular, geometric deep learning on graphs~(\mbox{GDLG})~\cite{gcnn,srmgcnn} has been proposed to solve the GMC problem, showing better performance than existing GMC methods including graph regularization based methods. To the best of our knowledge, there exists only one GDLG method for GMC~\footnote{Although GC-MC~\cite{gcmc} is a GDLG based method for MC, it is not for GMC because it only models the bi-partite rating matrix. Hence, the setting of GC-MC is different from that in this paper.}, which is called \underline{r}ecurrent \underline{m}ulti-\underline{g}raph \underline{c}onvolutional \underline{n}eural \underline{n}etwork~(\mbox{RMGCNN})~\cite{srmgcnn}. Based on spectral graph convolution framework~\cite{gcnn}, RMGCNN defines two-dimensional graph convolutional filters to process multi-graphs. The graph embeddings extracted by the two-dimensional graph convolutional filters are fed into a Long Short-Term Memory~(LSTM) recurrent neural network~(RNN)~\cite{rnn} to perform diffusion process, which is actually a smooth feature transformation process. After that, the final embeddings are used to complete the matrix completion task. A factorized~(matrix factorization) version, called separable RMGCNN~(sRMGCNN), is also proposed in~\cite{srmgcnn} for efficiency improvement. RMGCNN combines graph convolutional network~(GCN) and recurrent neural network~(RNN) together for GMC. Experimental results in~\cite{srmgcnn} show that the GCN part and RNN part can improve the performance of matrix completion simultaneously. However, matrix completion with pure GCN, named MGCNN in~\cite{srmgcnn}, is shown to be worse than RMGCNN in experiments. 

In this paper, we propose a new \mbox{GMC} method, called \underline{c}onvolutional \underline{g}eometric \underline{m}atrix \underline{c}ompletion~(CGMC), for recommendation with graphs among users/items. CGMC is a pure GCN-based method. The contributions of CGMC are listed as follows:
\begin{itemize}
	\item In CGMC, a new graph convolutional network is designed, by taking only the first two terms of Chebyshev polynomials in spectral graph convolution~\cite{gcnn} and adopting weighted policy to control the contribution between self-connections and neighbors for graph embeddings.
	\item Because the roles of users in the rating matrix and user graph are different, the latent factors~(embeddings) to represent users for rating matrix and those for user graph should also have some difference, although the users are the same. Hence, in CGMC, a fully connected layer is added to the output of GCN to project the user graph embeddings to a compatible space for rating matrix. Similar operations are also performed for items.
	\item CGMC integrates GCN and MC into a unified deep learning framework, in which the two components~(\mbox{GCN} and MC) can give feedback to each other.
	\item Experimental results on real datasets show that CGMC can outperform other state-of-the-art methods including RMGCNN. Hence, our work shows that pure GCN-based method can also achieve the best performance with properly designed deep architecture for graph convolution.
\end{itemize}

The following content is organized as follows. Section~\ref{sec:relatedWork} briefly discusses some related work. Section~\ref{sec:CGMF} presents the details of CGMC. Section~\ref{sec:experiment} shows experimental results. Section~\ref{sec:conclusion} concludes the paper.

\section{Related Work}\label{sec:relatedWork}
In this section, we introduce the related work of \mbox{CGMC}, including matrix completion~(MC), geometric matrix completion~(GMC), geometric deep learning on graphs~(GDLG), and GDLG based GMC.

\subsection{Matrix Completion}
Suppose $\bm{M}\in \mathbb{R}^{m\times n}$ is a rating matrix, with $m$ being the number of users and $n$ being the number of items. Given a subset of the entries $M_{ij},\,(i,\,j)\in\rm{\Omega}$, $|\Omega|\ll mn$. Matrix completion problem aims to estimate $M_{ij},\,\forall(i,\,j)\notin\Omega$. It is formulated as follows \cite{mc, mc1, mc2}:
\begin{equation} \label{eq:mc}
\min\limits_{\bm{Z}} \lVert \mathcal{P}_{\Omega} \bm{(M - Z)}\rVert^2_F + \gamma_z\lVert \bm{Z}\rVert_\star,
\end{equation}
where $\lVert \bm{Z}\rVert_\star$ is the nuclear norm of the matrix $\bm{Z}$. $ \mathcal{P}_{\Omega}$ is the projection operator, where $ [\mathcal{P}_{\Omega}(\bm{M})]_{ij}=M_{ij}$ if $(i,\,j)\in\Omega$, else $ [\mathcal{P}_{\Omega}(\bm{M})]_{ij}=0$.

One solution to the MC problem is to reformulate it as the following matrix factorization~(MF) problem \cite{mf,mf1,cabral13}:
\begin{align}\label{eq:mf}
\min\limits_{\bm{W,\,H}} \lVert \mathcal{P}_{\Omega} \bm{(M - WH^\top)}\rVert^2_F + \frac{\gamma_w}{2}\lVert \bm{W}\rVert^2_F + \frac{\gamma_h}{2}\lVert \bm{H}\rVert^2_F,
\end{align}
where $\bm{W}$ and $\bm{H}$ are latent factor representation for users and items, respectively.

\subsection{Geometric Matrix Completion}
Geometric matrix completion~(GMC)~\cite{ gmc2, gmc1, gmc3, grals} has been developed to exploit the relationship~(link) graph among users/items to assist the matrix completion process. One kind of GMC methods is to adopt graph Laplacian for regularization. GRALS~\cite{grals} is one representative of this kind, which is formulated as follows:
\begin{align}\label{grals}
\min\limits_{\bm{W,\,H}} &\lVert \mathcal{P}_{\Omega} \bm{(M - WH^\top)}\rVert^2_F + \frac{\gamma_w}{2}\lVert \bm{W}\rVert^2_F + \frac{\gamma_h}{2}\lVert \bm{H}\rVert^2_F \nonumber \\
& +\frac{\beta_w}{2}\mathrm{tr}(\bm{W}^\top \bm{L}_W \bm{W}) + \frac{\beta_h}{2}\mathrm{tr}(\bm{H}^\top \bm{L}_H \bm{H}),
\end{align}
where $\bm{L}_W$ and $\bm{L}_H$ are the normalized graph Laplacian of user graph $\bm{A}$ and item graph $\bm{B}$, respectively. $\bm{L}_W = \bm{I} - \bm{D}_W^{-\frac{1}{2}}\bm{A}\bm{D}_W^{-\frac{1}{2}}$ and $\bm{D}_W$ is a diagonal matrix with diagonal entry $[D_W]_{ii} = \sum_{j} A_{ij}$. $\bm{I}$ is an identity matrix whose dimensionality depends on the context. $\bm{L}_H$ can be similarly computed based on $\bm{B}$.

\subsection{Geometric Deep Learning on Graphs}
Recently, there have appeared a few works that attempt to perform geometric deep learning on graphs~(GDLG)~\cite{gdnn1, gdnn2, gdnn3,Ying18,Gao18}. In particular, inspired by spectral graph theory in graph signal processing~\cite{spectral2,spectral1}, spectral graph convolution is proposed in~\cite{gdnn5,gdnn4}.

According to~\cite{gcnn}, an efficient graph convolution operation is proposed as follows:
\begin{align} \label{eq:spectralGC}
g_{\bm{\theta}} \star \bm{x} &\approx \sum\nolimits_{k=0}^{K} \theta_k T_k(\bm{\tilde{L}})\bm{x},
\end{align}
where $\bm{x}\in\mathbb{R}^{N}$ is an input signal of a graph $\bm{G}$~($N$ is the number of nodes in graph), $\bm{\theta}$ is a learnable filter, $\bm{\tilde{L}} = \frac{2}{\lambda_{max}} \bm{L} - \bm{I}$, $\lambda_{max}$ denotes the largest eigenvalue, $\bm{L}$ denotes the symmetric graph Laplacian matrix of graph $G$, $T_0(\bm{\tilde{L}}) = \bm{I}$, $T_1(\bm{\tilde{L}}) = \bm{\tilde{L}}$, $T_k(\bm{\tilde{L}}) = 2\bm{\tilde{L}}\odot T_{k-1}(\bm{\tilde{L}}) - T_{k-2}(\bm{\tilde{L}})$, $\odot$ is Hadamard product.

A simplified variant of spectral graph convolutional network~(GCN) is proposed in ~\cite{gcn}. We call it \mbox{GCN-kw} in this paper. By assuming $K=1$, $\lambda_{max}\approx2$ and with imposed constraints and renormalization, GCN-kw is formulated as follow:
\begin{gather}
\begin{aligned}
g_{\theta} \star \bm{x} = \theta \bm{\tilde{D}}^{-\frac{1}{2}}\bm{\tilde{G}}\bm{\tilde{D}}^{-\frac{1}{2}}\bm{x},
\end{aligned}
\end{gather}
where $\bm{\tilde{G}} = \bm{G}+\bm{I}$ and $\tilde{D}_{ii} = \sum_{j}\tilde{G}_{ij}$. Here, we can see that self-connections and neighbors contribute equally to graph embeddings, which is not flexible enough.

\subsection{GDLG based GMC}
To the best of our knowledge, RMGCNN~\cite{srmgcnn} is the only work which has applied geometric deep learning on graphs~(GDLG) for GMC. RMGCNN adopts GCN~\cite{gcnn} to extract graph embeddings for users and items, and then combines with recurrent neural network~(RNN) to perform diffusion process, which is actually a smooth feature transformation process. The factorized version of RMGCNN~\cite{srmgcnn} is shown as follows:
\begin{gather}
\begin{aligned}
\min\limits_{\bm{\theta}_r,\bm{\theta}_c, \bm{\theta}_{rnn}}
\quad\lVert\mathcal{P}_{\Omega}(\bm{M} - \bm{W}_{\bm{\theta}_r,\bm{\theta}_{rnn}}^{(T)}(\bm{H}_{\bm{\theta}_c,\bm{\theta}_{rnn}}^{(T)})^\top)\rVert^2_F
+ \frac{\mu}{2}(\lVert\bm{W}_{\bm{\theta}_r,\bm{\theta}_{rnn}}^{(T)}\rVert_{\mathcal{G}_r} + \lVert\bm{H}_{\bm{\theta}_c,\bm{\theta}_{rnn}}^{(T)}\rVert_{\mathcal{G}_c})
\end{aligned}
\end{gather}
where $\bm{\theta}_r$ and $\bm{\theta}_c$ denotes the parameter of GCN, $\bm{\theta}_{rnn}$ denotes the parameter of RNN, $\bm{W}_{\bm{\theta}_r,\bm{\theta}_{rnn}}^{(T)}$ and $\bm{H}_{\bm{\theta}_c,\bm{\theta}_{rnn}}^{(T)}$ are the graph embeddings extracted by GCN and RNN for users and items respectively, $\mathcal{G}_r$ and $\mathcal{G}_c$ are graphs on users and items respectively, $T$ denotes the graph embedding iterates for $T$ iterations, $\lVert\cdot\rVert_{\mathcal{G}_r}$ and $\lVert\cdot\rVert_{\mathcal{G}_c}$ represent graph Laplacian regularization.

\section{Convolutional GMC}\label{sec:CGMF}
In this section, we present the details of our new GDLG-based GMC method, called \underline{c}onvolutional \underline{g}eometric \underline{m}atrix \underline{c}ompletion~(CGMC). CGMC is a pure GCN-based method. CGMC shows that GMC with only GCN can outperform the GCN+RNN method RMGCNN to achieve the state-of-the-art performance.

CGMC is formulated as follows. Firstly, a new GCN is proposed to extract graph embedding, which is called \emph{convolutional graph embedding}~(CGE) in this paper, for user/item representation. Then, a fully connected layer is added to the output of GCN to project user/item graph embeddings to a compatible space for rating matrix. After that, GCN and MC are integrated into a unified deep learning framework to get CGMC. 

\subsection{Convolutional Graph Embedding~(CGE)}
Here, we propose a new GCN to get the convolutional graph embedding~(CGE) for graph node representation. 

By taking $K=1$ in the spectral graph convolution of~(\ref{eq:spectralGC}), we have:
\begin{gather*}
\begin{aligned}
g_{\bm{\theta}} \star \bm{x} = \theta_0 \bm{x} + \theta_1 \bm{\tilde{L}}\bm{x} 
= \big(\big(\theta_0 + \theta_1 (\frac{2}{\lambda_{max}} - 1)\big)\bm{I} - \theta_1 \frac{2}{\lambda_{max}}\bm{S}\big)\bm{x},
\end{aligned}
\end{gather*}
where we let $\bm{S} = \bm{D}^{-\frac{1}{2}}\bm{G}\bm{D}^{-\frac{1}{2}}$ and $\bm{G}$ is the link matrix of the graph with $N$ nodes. Since $\theta_0, \, \theta_1$ are free parameters, and there is no constraints between the coefficients of $\bm{I}$ and $\bm{S}$, we can let $\alpha_0 = \theta_0 + \theta_1 (\frac{2}{\lambda_{max}} - 1), \, \alpha_1 = -\theta_1 \frac{2}{\lambda_{max}}$. Furthermore, $\alpha_0,\, \alpha_1$ are still free parameters. We let $\alpha_0 = \theta\sigma_0$, $\alpha_1 = \theta\sigma_1$, and get
\begin{gather*}
\begin{aligned}
g_{\bm{\theta}} \star \bm{x} &= \theta(\sigma_0\bm{I} + \sigma_1 \bm{S})\bm{x},
\end{aligned}
\end{gather*}
Here, $\sigma_0$ and $\sigma_1$ can be explained as a weight controlling the contribution between self-connections and neighbors. We constrain $\sigma_0+ \sigma_1=1$ and $\sigma_0,\, \sigma_1 \in (0,1)$. For convenience, we denote $\sigma = \sigma_0$, and get
\begin{gather}\label{eq:gconv1}
\begin{aligned}
g_{\bm{\theta}} \star \bm{x} &= \theta(\sigma\bm{I} + (1-\sigma) \bm{S})\bm{x},
\end{aligned}
\end{gather}
The eigenvalues of $\sigma\bm{I} + (1-\sigma) \bm{S}$ are in $[-1, \,1]$, which can be easily verified according to Lemma 1.7 in~\cite{spectralgraphtheory}. Hence, such a filter won't result into numerically unstable outputs. 

Generally, proportion of contribution between self-connections and neighbors should be different for different nodes. Hence, we further proposed to utilize an independent learnable parameter to control the contribution between self-connetions and neighbors, which is formulated as follows:
\begin{gather}\label{eq:gconv2}
\begin{aligned}
g_{\bm{\theta}} \star \bm{x} &= \theta(\mathrm{diag}(\bm{\sigma}) + (\bm{I}-\mathrm{diag}(\bm{\sigma})) \bm{S})\bm{x},
\end{aligned}
\end{gather}
where $\bm{\sigma}\in\mathbb{R}^{N}$ and $\bm{\sigma}\in(0,1)^N$, $\mathrm{diag}(\bm{\sigma})$ is a diagonal matrix. In the following Theorem, we will show that the eigenvalues of $\mathrm{diag}(\bm{\sigma}) + (\bm{I}-\mathrm{diag}(\bm{\sigma})) \bm{S}$ are in $[-1,1]$. 
\begin{theorem}\label{theo1}
	For the diagonal matrix $\bm{Z} = \mathrm{diag}(\bm{\sigma}) \in \mathbb{R}^{N\times N}$ and the symmetric matrix $\bm{S} \in \mathbb{R}^{N\times N}$, if $Z_{ii}\in(0,1)$, and the eigenvalues of $\bm{S}$, denoted as $\lambda_1,\cdots,\lambda_N$, are all bounded in the range $[-1,1]$, then eigenvalues of $\bm{Q} = \bm{Z} + (\bm{I} - \bm{Z})\bm{S}$ are all bounded in the range $[-1,1]$.
\end{theorem}
Hence, such a filter won't result into numerically unstable outputs either. We can see that Equation~(\ref{eq:gconv2}) is more general than Equation~(\ref{eq:gconv1}), and Equation~(\ref{eq:gconv1}) is a special case of Equation~(\ref{eq:gconv2}) by setting the entries of $\bm{\sigma}$ to be the same. The proof of Theorem~\ref{theo1} is in the Appendix.

When the input signal is multi-dimensional, denoted by $\bm{V} \in \mathbb{R}^{N\times r}$ with $N$ being the number of nodes and $r$ being the dimensionality, we can get the formulation of multi-dimensional graph convolution as follows. We use $\bm{v}_j$ to denote the $j$th column of $\bm{V}$, which is the $j$th input signal.
\begin{align*}
\bm{\hat{v}}_i = \sum_{j=1}^{r} \Theta_{ji}\big(\mathrm{diag}(\bm{\sigma}) + (1-\mathrm{diag}(\bm{\sigma}))\bm{S}\big)\bm{v}_j,
\end{align*}
where $i\in\{1,\,\cdots,\,q\}$, $q$ is the dimensionality of the output signal, $\Theta_{ji}$ is the filter parameter of the $i$-th output signal defined on the $j$-th input signal, $\bm{\Theta}\in \mathbb{R}^{r\times q}$. Then we can get,
\begin{align*}
\bm{\hat{V}} = (\mathrm{diag}(\bm{\sigma}) + (1-\mathrm{diag}(\bm{\sigma}))\bm{S})\bm{V}\bm{\Theta},
\end{align*}
which transforms the node representation from $\bm{V}\in\mathbb{R}^{N\times r}$ to $\bm{\hat{V}}\in\mathbb{R}^{N\times q}$ through one-layer graph convolution with the convolution parameter $\bm{\Theta}$.

By stacking the above formulation to multiple layers, we can get a deep model for CGE:
\begin{gather}
\begin{aligned}\label{cge-1}
\bm{V}^{(\ell+1)} = f\big((\mathrm{diag}(\bm{\sigma}) + (1-\mathrm{diag}(\bm{\sigma}))\bm{S})\bm{V}^{(\ell)}\bm{\Theta}^{(\ell)}\big),
\end{aligned}
\end{gather}
where $\bm{V}^{(\ell)}$ is the output signal of the $\ell$-th layer, $\bm{\Theta}^{(\ell)}$ is the convolution parameter of the $\ell$-th layer, and $f(\cdot)$ is an activation function.

\subsection{Model of CGMC}
Our CGMC can also be used for the nuclear norm regularization formulation in~(\ref{eq:mc}), by adopting similar techniques in RMGCNN~\cite{srmgcnn}. However, as pointed out by~\cite{srmgcnn}, the nuclear norm regularization formulation has high storage consumption, which is not feasible for large-scale dataset. Hence, in this paper, we adopt the MF formulation in~(\ref{eq:mf}) for our CGMC.

Suppose $\bm{X}\in \mathbb{R}^{m\times r_m}$ denotes the input user features, $\bm{Y}\in \mathbb{R}^{n\times r_n}$ denotes the input item features, with $m$ and $n$ being the number of users and items respectively, $r_m$ and $r_n$ being the feature dimensionality for users and items respectively. If $\bm{X}$ or $\bm{Y}$ is not available, we set $\bm{X} = \bm{I}$ or $\bm{Y} = \bm{I}$. $\bm{A}$ and $\bm{B}$ are user graph and item graph. Then the CGE for users and items can be generated by applying (\ref{cge-1}) to graph $\bm{A}$ and $\bm{B}$:
\begin{gather}\label{cgmc-1}
\begin{aligned}
&\bm{\hat{L}}_W = \mathrm{diag}(\bm{\sigma}_W) + (1-\mathrm{diag}(\bm{\sigma}_W))\bm{D}^{-\frac{1}{2}}_W\bm{A}\bm{D}^{-\frac{1}{2}}_W,\,\bm{X}^{(\ell + 1)} = f(\bm{\hat{L}}_W\bm{X}^{(\ell)}\bm{\Theta}^{(\ell)}_W)\\
&\bm{\hat{L}}_H = \mathrm{diag}(\bm{\sigma}_H) + (1-\mathrm{diag}(\bm{\sigma}_H))\bm{D}^{-\frac{1}{2}}_H\bm{B}\bm{D}^{-\frac{1}{2}}_H,\,\bm{Y}^{(\ell + 1)} = f(\bm{\hat{L}}_H\bm{Y}^{(\ell)}\bm{\Theta}^{(\ell)}_H),
\end{aligned}
\end{gather}
where $\bm{\sigma}_W\in(0,1)^m$, $\bm{\sigma}_H\in(0,1)^n$, $\bm{D}_W$ and $\bm{D}_H$ are diagonal degree matrices of $\bm{A}$ and $\bm{B}$ respectively, $\bm{X}^{(\ell)}$ and $\bm{Y}^{(\ell)}$ are the output feature representation of the $\ell$-th layer, $\bm{X}^{(0)} = \bm{X}$ and $\bm{Y}^{(0)} = \bm{Y}$, $f(\cdot)$ is an activation function, here we take $f(\cdot)=\tanh(\cdot)$, $\bm{\Theta}^{(\ell)}_W$ and $\bm{\Theta}^{(\ell)}_H$ are convolution parameters which play the same role as $\bm{\Theta^{(\ell)}}$ in~(\ref{cge-1}). $\bm{\Theta}^{(0)}_W\in\mathbb{R}^{r_m\times d}$, $\bm{\Theta}^{(\ell)}_W\in\mathbb{R}^{d\times d}$ ($\ell > 0$), $\bm{\Theta}^{(0)}_H\in\mathbb{R}^{r_n\times d}$ and $\bm{\Theta}^{(\ell)}_H\in\mathbb{R}^{d\times d}$ ($\ell > 0$).

\textbf{Fully-Connected Layer after CGE}
For a specific user, he/she plays a role in the user graph, and he/she also plays another role in the rating matrix. These two roles are different. Intuitively, the latent factors~(embeddings) to represent these two different roles of this user should also have some difference. Items also have similar property.

To capture the difference between these two roles, a fully connected layer is added to the output of GCN to project the CGE to a compatible space for rating matrix. The formulation is as follows:
\begin{gather}\label{proj}
\begin{aligned}
\bm{\hat{W}} = f(\bm{X}^{(L)}\bm{\Theta}^{(L)}_W + \bm{1}\bm{b}^\top_W),\,
\bm{\hat{H}} = f(\bm{Y}^{(L)}\bm{\Theta}^{(L)}_H + \bm{1}\bm{b}^\top_H),
\end{aligned}
\end{gather}
where $\bm{X}^{(L)}$ and $\bm{Y}^{(L)}$ are the output user features and item features of the CGE with $L$ layers, $\{\bm{\Theta}^{(L)}_W,\bm{b}_W\}$ and $\{\bm{\Theta}^{(L)}_H, \bm{b}_H\}$ are parameters of the fully connected layer for user CGE and item CGE, $f(\cdot) = \tanh(\cdot)$. $\bm{\Theta}^{(L)}_W\in\mathbb{R}^{d\times d}$ and $\bm{\Theta}^{(L)}_H\in\mathbb{R}^{d\times d}$.

This is one key difference between our method and other methods like RMGCNN. In our experiments, we will verify that this fully connected layer will improve the performance of CGE.

\textbf{Objective Function} By applying CGE and with the projection by the fully connected layer, CGMC is formulated as follows:
\begin{gather}\label{eq:CGMCobj1}
\min\limits_{\bm{\mathcal{W},\mathcal{H}}}\quad\lVert\mathcal{P}_{\Omega}(\bm{M} - \bm{\hat{W}}\bm{\hat{H}}^\top)\rVert^2_F + \frac{\gamma}{2} \sum\nolimits_{\ell=0}^{L} (\lVert \bm{\Theta}^{(\ell)}_W \rVert^2_F + \lVert \bm{\Theta}^{(\ell)}_H\rVert^2_F),
\end{gather}
where $\bm{\mathcal{W}}$ denotes $\{\bm{\Theta}^{(0)}_W,\, \bm{\Theta}^{(1)}_W, \,\cdots,\,\bm{\Theta}^{(L)}_W,\,\bm{b}_W\}$ and $\bm{\mathcal{H}}$ denotes $\{\bm{\Theta}^{(0)}_H,\, \bm{\Theta}^{(1)}_H, \,\cdots,\,\bm{\Theta}^{(L)}_H,\,\bm{b}_H\}$. From~(\ref{eq:CGMCobj1}), it is easy to find that CGMC seamlessly integrates GCN and MC into a unified deep learning framework, in which GCN and MC can give feedback to each other for performance improvement.
\subsection{Learning}
We adopt mini-batch gradient descent with momentum~\cite{momentum} to optimize the parameters $\bm{\mathcal{W}}$, $\bm{\mathcal{H}}$, $\bm{\sigma}_W$ and $\bm{\sigma}_H$. For CGMC, we do not directly optimize $\bm{\sigma}_W$ and $\bm{\sigma}_H$. Since $\bm{\sigma}_W$ and $\bm{\sigma}_H$ are constrained in range $(0,1)$, we can get rid of the constraints by learning new parameters: $\bm{\sigma}_W = 1/(1 + \exp(-\bm{\sigma}^\prime_W))$, $\bm{\sigma}_H = 1/(1 + \exp(-\bm{\sigma}^\prime_H))$. Hence, CGMC can be reformulated as follows:
\begin{gather}\label{eq:CGMCobj3}
\begin{aligned}
\min\limits_{\bm{\mathcal{W},\mathcal{H}},\bm{\sigma}^\prime_W, \bm{\sigma}^\prime_H}\quad&\lVert\mathcal{P}_{\Omega}(\bm{M} - \bm{\hat{W}}\bm{\hat{H}}^\top)\rVert^2_F +  \frac{\gamma}{2} \mathcal{L}_{reg}.
\end{aligned}
\end{gather}

The entire learning procedure for CGMC is summarized in Algorithm~\ref{alg:CGMC}. Firstly, for each mini-batch training iteration, we sample a batch user-item pairs. Secondly, in the process of forward propagation, we perform the defined graph convolution operation to get the embeddings for the sampled users and items according to~(\ref{cgmc-1}). And then, we project the convolutional graph embeddings of user and item to a compatible space according to~(\ref{proj}). Finally, in the process of back propagation, we update the parameters.

\subsection{Comparison to Related Work}
\begin{wrapfigure}{r}{0.6\linewidth}
	\begin{minipage}{1.0\linewidth}
		\vskip -0.35in
		\begin{algorithm}[H]
			\caption{Learning Algorithm of CGMC}
			\label{alg:CGMC}
			\begin{algorithmic}
				\State {\bfseries Input:} $\bm{M}$, $\bm{A}$, $\bm{B}$, $\bm{X}$, $\bm{Y}$, $L$, $\eta$, $\gamma$, batchsize.
				\State {\bfseries Preprocess:} Initialize $\bm{\mathcal{W}}$, $\bm{\mathcal{H}}$, $\bm{\sigma}_W^\prime$, $\bm{\sigma}_H^\prime$, $\bm{\hat{L}}_W$ and $\bm{\hat{L}}_H$.
				\For{$outter=1$ {\bfseries to} $T$}
				\For{$inner=1$ {\bfseries to} minibatch}
				\State Sample a batch user-item pairs from training data.
				\State \emph{Forward Propagation}:
				\State \quad Calculate $\bm{X}^{(L)}$, $\bm{Y}^{(L)}$ according to (\ref{cgmc-1}).
				\State \quad Project $\bm{X}^{(L)}$, $\bm{Y}^{(L)}$ to $\bm{\hat{W}}$, $\bm{\hat{H}}$ according to (\ref{proj}).
				\State \emph{Back Propagation}:
				\State \quad Calculate the gradients of parameters.
				\State \quad Update $\bm{\mathcal{W}}$: $\bm{\mathcal{W}} \leftarrow \bm{\mathcal{W}} + \eta \bigtriangledown_{\bm{\mathcal{W}}}$.
				\State \quad Update $\bm{\mathcal{H}}$: $\bm{\mathcal{H}} \leftarrow \bm{\mathcal{H}} + \eta \bigtriangledown_{\bm{\mathcal{H}}}$.
				\State Repeat the above procedure to update $\bm{\mathcal{H}}$ and $\bm{\sigma}_H^\prime$.
				\EndFor
				\EndFor
			\end{algorithmic}
		\end{algorithm}
	\end{minipage}
	\vskip -0.2in
\end{wrapfigure}
The most related work to our CGMC is RMGCNN~(sRMGCNN)~\cite{srmgcnn} and GCN-kw in~\cite{gcn}. Here we discuss the difference between them and our CGMC.

As mentioned above, sRMGCNN is a factorized~(MF) version of RMGCNN. Because we only focus on the factorized version in this paper due to its efficiency, RMGCNN in this paper refers to sRMGCNN unless otherwise stated. CGMC is different from RMGCNN in the following aspects. Firstly, CGMC adopts a different GCN to extract graph embeddings, and the newly designed GCN in CGMC is better than that in RMGCNN which will be verified in experiments. Secondly, RMGCNN adopts both GCN and RNN for GMC, while our CGMC adopts only GCN without RNN. Thirdly, a fully connected layer is introduced in our CGMC for space compatibility.

CGMC is different from GCN-kw in the following aspects. Firstly, \mbox{GCN-kw} is proposed for semi-supervised learning, and it has not been used for MC. Secondly, the GCN in CGMC adopts weighted policy to control the contribution between self-connections and neighbors for graph embedding, while the self-connections and neighbors in GCN-kw contribute equally for graph embedding. Hence, the GCN in CGMC is more flexible than GCN-kw. Thirdly, the filter of GCN-kw~\cite{gcn} will result into numerically unstable outputs if no further operation is performed, while the filter of GCN in CGMC will not. Although GCN-kw is not proposed for GMC, we adapt it for GMC in this paper and find that CGMC achieves better performance than GCN-kw based method in our experiment.

\section{Experiment}\label{sec:experiment}
We evaluate our proposed CGMC and other baselines on collaborative filtering datasets. Our implementation is based on PyTorch~\cite{paszke17} with a NVIDIA TitanXP GPU server. 
\begin{table}[t]
	\vskip -0.1in
	\caption{Statistics of datasets for evaluation. For the "Graphs", `Users/Items' denotes that both user graph and item graph are used, `Users' denotes only user graph is used, and `Items' denotes only item graph is used.}
	\label{statistic}
	\begin{center}
		\begin{small}
			\begin{tabular}{lcclcccr}
				\toprule
				Dataset & \#Users & \#Items & Graphs & \#Ratings & Density & Rating levels \\
				\midrule
				ML-100K    & 943 & 1682 & Users/Items & 100,000 & 0.0630 & $1,\,2,\,\cdots,\,5$\\
				Douban & 3000 & 3000 & Users & 136,891 & 0.0152 & $1,\,2,\,\cdots,\,5$\\
				Flixster    & 3000 & 3000 & Users/Items & 26,173 & 0.0029 & $0.5,\,1,\,\cdots,\,5$\\
				YahooMusic    & 3000 & 3000 & Items & 5,335 & 0.0006 & $1,\,2,\,\cdots,\,100$\\
				ML-1M	& 6,040 & 3,706 & Users/Items & 1,000,209 & 0.0447 & $1,\,2,\,\cdots,\,5$\\
				ML-10M	& 69,878 & 10,677 & Items & 10,000,054 & 0.0134 & $0.5,\,1,\,\cdots,\,5$\\	
				\bottomrule
			\end{tabular}
		\end{small}
	\end{center}
	\vskip -0.2in
\end{table}
\subsection{Datasets}
Firstly, as in RMGCNN~\cite{srmgcnn}, we evaluate \mbox{CGMC} and other baselines on four small datasets: Movielens-100K
~\footnote{https://grouplens.org/datasets/movielens/}
~(ML-100K), Douban, Flixster, YahooMusic. For fair comparison, we use the same training/test data partition that are provided by~\cite{srmgcnn}
~\footnote{https://github.com/fmonti/mgcnn}
. Specifically, in RMGCNN, graph of ML-100K is constructed with user/item feature, graph of Douban is a social graph, graph of Flixster is constructed with original rating matrix, graph of YahooMusic is constructed with item feature. Secondly, as in GC-MC~\cite{gcmc}, we evaluate \mbox{CGMC} and other baselines on two large datasets: Movielens-1M~(ML-1M), Movielens-10M~(ML-10M). For fair comparison, we use the same training/test data partition that are provided by~\cite{gcmc}
. Since there are no public graphs provided for ML-1M and ML-10M, we construct user/item graphs via 10-nearest neighbors measured by Euclidean distance of features. Statistics of datasets are summarized in Table~\ref{statistic}.

\subsection{Baselines and Settings}
	\begin{wraptable}{r}{0.3\linewidth}
	\begin{center}
		\vskip -0.26in
		\caption{Performance (RMSE) on ML-100K. The results of baselines are from~\cite{srmgcnn} and~\cite{gcmc}.}
		\vskip -0.05in
		\label{result1}
		\begin{small}
			\begin{tabular}{lcr}
				\toprule
				Method & RMSE\\
				\midrule
				Global Mean   & 1.154\\
				User Mean & 1.063\\
				Movie Mean    & 1.033\\
				MC \cite{mc1}    & 0.973\\
				IMC \cite{imc1, imc2}& 1.653\\
				GMC \cite{gmc} & 0.996\\
				GRALS \cite{grals} & 0.945\\
				GC-MC \cite{gcmc} & 0.905\\
				RMGCNN \cite{srmgcnn} & 0.929\\
				\midrule
				\textbf{CGMC}& \textbf{0.893}\\
				\bottomrule
			\end{tabular}
		\end{small}
	\end{center}
	\vskip -0.15in
\end{wraptable}
\textbf{Baselines} We compare CGMC with baselines such as MC \cite{mc1}, IMC \cite{imc1, imc2}, GMC \cite{gmc}, GRALS\cite{grals}, RMGCNN \cite{srmgcnn}, GC-MC~\cite{gcmc}. MC learns the full matrix with a nuclear norm regularization. IMC utilizes the features of users and items to formulate an inductive matrix model for approximating the target. GMC learns a full matrix that approximates the observed rating matrix and constrains the full matrix by applying graph Laplacian regularization on it. GRALS learns the factorized matrices of the target by applying graph Laplacian regularization on the factorized matrices. RMGCNN approximates the full matrix with the dot product between user and item embeddings, which are generated from two-dimensional graph convolutional filters followed by LSTM-RNN. GC-MC defines graph convolution operation on bi-partie graph, which is constructed from rating matrix by discretizing ratings into different levels, and learns the full matrix by reformulating the learning problem as a link prediction problem. Since RMGCNN only supports full-batch training due to the graph laplacian regularization on the full rating matrix, it cannot fit into the GPU memory for ML-1M and ML-10M. To compare CGMC with RMGCNN on ML-1M and ML-10M, we remove the graph laplacian regularization on the full rating matrix to enable RMGCNN to train in mini-batch. Hence, the graph information is only utilized in the GCN and RNN part to extract graph embeddings for users and items.

\textbf{Settings} Since the graphs of ML-100K, YahooMusic, ML-1M and ML-10M are contructed from features, and features of Douban and Flixster are unavailable, we implement featureless version of CGMC in all our experiments, where we set $\bm{X},\, \bm{Y} = \bm{I}$. For each dataset, we randomly sample instances from training set as validation set that has the same number as test set. We repeat the experiments 5 times and report the mean of results. On all datasets, we adopt a version of single graph convolution layer for CGMC to compare with baselines. The optimization algorithm we use is mini-batch gradient descent with momentum~\cite{momentum}. The regularization parameter $\gamma$ is selected from $[10^{-8},\,10^8]$. We use validation set to tune this hyper-parameter. Other hyper-parameters can refer to the Appendix. As in RMGCNN~\cite{srmgcnn}, root mean square error~(RMSE) is adopted as metric for evaluation. The smaller the RMSE is, the better the performance will be. All methods run with best hyper-parameters tuned with validation set on different datasets. The standard deviation is very small (approximately 0.001 for small datasets, 0001 for large datasets), we omit it in the tables.

%

\subsection{Result}
\begin{wraptable}{r}{0.6\textwidth}
		\begin{center}
			\vskip -0.25in
			\caption{Performance~(RMSE) on Douban, Flixster and YahooMusic. Flixster-U only uses user graph. MC is implemented by ourself. Results of other baselines are from~\cite{gcmc}.}
			\vskip -0.05in
			\label{result2}
			\begin{small}
				\begin{tabular}{lccccc}
					\toprule
					Method & Douban & Flixster & Flixster-U &YahooMusic\\
					\midrule
					MC	& 0.845 & 1.533 & 1.534 & 52.0\\
					GRALS  & 0.833 & 1.313 & 1.243 & 38.0\\
					GC-MC & 0.734 & 0.917 & 0.941 & 20.5\\
					RMGCNN & 0.801 & 1.179 & 0.926 & 22.4\\
					\midrule
					\textbf{CGMC} & \textbf{0.728} & \textbf{0.878} & \textbf{0.900} & \textbf{18.9}\\
					\bottomrule
				\end{tabular}
			\end{small}
		\end{center}
	\vskip -0.2in
\end{wraptable}
The results on ML-100K are reported in Table~\ref{result1}, where the results of baselines are from~\cite{srmgcnn}. Because the training/test data partition of this paper and graphs are exactly the same as that in~\cite{srmgcnn}, the comparison is fair. From Table~\ref{result1}, we can find that our CGMC outperforms all the other baselines, including graph regularization methods and GDLG-based methods, to achieve the best performance.

The results on Douban, Flixster and YahooMusic are reported in Table~\ref{result2}. Once again, we can find that CGMC outperforms other state-of-the-art baselines to achieve the best performance. 

\begin{wraptable}{r}{0.52\textwidth}
	\begin{center}
		\vskip -0.25in
		\caption{Performance~(RMSE) on ML-1M and ML-10M. ML-1M and ML-10M have five train/test splits, the mean RMSE on the five splits are reported. Numbers in $(\cdot)$ denote the training time per epoch.}
		\label{result3}
		\begin{small}
			\begin{tabular}{lccccc}
				\toprule
				Method & ML-1M & ML-10M\\
				\midrule
				GC-MC & 0.8318 (202s)  & 0.7772 (3116s)\\
				RMGCNN & 0.8653 (212s) & 0.8329 (508s)\\
				\midrule
				\textbf{CGMC} & \textbf{0.8275} (7s) & \textbf{0.7754} (43s)\\
				\bottomrule
			\end{tabular}
		\end{small}
	\end{center}
	\vskip -0.2in
\end{wraptable}
The results on ML-1M and ML-10M are reported in Table~\ref{result3}. For fair comparision, we set batchsize to be the same for all baselines. GC-MC is trained with the code provided by the corresponding authors. RMGCNN is implemented according to the code provided by the corresponding authors. We can find that CGMC outperforms other baselines in terms of both accuracy and speed, and it just verifies the scalability of CGMC. 

\subsection{Effect of Fully-Connected Layer}
To demonstrate the effectiveness of the fully-connected layer in GCN proposed by us, we remove the fully-connected layer after CGE. The CGMC variant without fully connected layer is denoted as CGMC-0, and CGMC is with fully-connected layer. We compare CGMC-0 to CGMC under the conditions where GCN grows from 1 layer to 4 layers. 

The results are shown in Table~\ref{result4}. From Table~\ref{result4}, we can observe that with different number of layers for GCN, the improvements of CGMC over CGMC-0 are significant. These results verify the effectiveness of the fully-connected layer in CGMC.
\begin{table}[h]
	\begin{center}
		\vskip -0.1in
		\caption{The influence of the fully-connected layer. CGMC-0 denotes CGMC without fully connected layer, and CGMC denotes CGMC with fully connected layer.}
		\label{result4}
		\begin{small}
			\begin{tabular}{lccccr}
				\toprule
				Method & ML-100K & Douban & Flixster & YahooMusic\\
				\midrule
				CGMC-0/CGMC (1 layer) & 0.957/\textbf{0.893} & 0.958/\textbf{0.728} & 1.083/ \textbf{0.878} & 37.0/\textbf{18.9}\\
				CGMC-0/CGMC (2 layers) & 0.906/\textbf{0.897} & 0.742/\textbf{0.735} & 0.902/\textbf{0.880} & 36.8/\textbf{19.6}\\
				CGMC-0/CGMC (3 layers) & 0.913/\textbf{0.906} & 0.757/\textbf{0.740} & 0.910/\textbf{0.880} & 36.8/\textbf{19.8}\\
				CGMC-0/CGMC (4 layers) & 0.917/\textbf{0.904} & 0.765/\textbf{0.746} & 0.921/\textbf{0.886} & 36.8/\textbf{19.9}\\
				\bottomrule
			\end{tabular}
		\end{small}
	\end{center}
\vskip -0.1in
\end{table}

\subsection{Effect of Weighted Policy in GCN}
To demonstrate the effectiveness of the weighted policy proposed in our newly designed GCN, we replace our GCN in CGMC by the GCN-kw~\cite{gcn}~\footnote{GCN-kw is not proposed for GMC, we adapt it for GMC here.}. The resulting model is denoted as GMC-GCN-kw. We design two variants of GMC-GCN-kw. GMC-GCN-kw denotes the variant of our CGMC by only replacing our GCN by GCN-kw, with all other parts fixed. It means that GMC-GCN-kw also includes a fully connected layer which is proposed by us. GMC-GCN-kw-0 denotes a variant of GMC-GCN-kw without the fully connected layer. 

The results are reported in Table~\ref{result4}. We can observe that \mbox{CGMC} performs better than \mbox{GMC-GCN-kw} on all datasets. The results show the effectiveness of adopting weighted policy to control the contribution between self-connections and neighbors for graph embeddings. Compared with \mbox{GMC-GCN-kw-0}, the performance improvement of \mbox{GMC-GCN-kw} once again verifies the effectiveness of the fully connected layer proposed by us.
\begin{table}[h]
	\begin{center}
		\vskip -0.2in
		\caption{The influence of our weighted policy in the GCN of CGMC.}
		\label{result5}
		\begin{small}
		\begin{tabular}{lcccccc}
			\toprule
			Method & ML-100K & Douban & Flixster & Flixster-U &YahooMusic\\
			\midrule
			GMC-GCN-kw-0 (1 layer) & 1.088 & 1.709 & 1.528 & 1.581 & 34.6\\
			GMC-GCN-kw-0 (2 layers) &1.049 & 0.755 & 0.925 & 1.112 & 34.4\\
			GMC-GCN-kw-0 (3 layers) & 1.076 & 0.773 & 0.972 & 1.158 & 34.5\\
			GMC-GCN-kw-0 (4 layers) & 1.082 & 0.779 & 1.005 & 1.157 & 34.5\\
			GMC-GCN-kw & 1.010 & 0.737 & 0.889 & 0.937 & 19.9\\
			\midrule
			\textbf{CGMC} & \textbf{0.893} & \textbf{0.728} & \textbf{0.875} & \textbf{0.900} & \textbf{18.9}\\
			\bottomrule
		\end{tabular}
	\end{small}
	\end{center}
\vskip -0.2in
\end{table}

\subsection{Sensitivity to Hyper-parameters}
In CGMC, $\gamma$ and embedding dimension $d$ are two important hyper-parameters. Here, we study the sensitivity of $\gamma$ on Flixster and YahooMusic, and the sensitivity of embedding dimension $d$ on two large datasets. 

\begin{figure}[h]
	\begin{minipage}{0.5\linewidth}	
		\begin{minipage}{0.49\linewidth}
			\begin{center}
				\centerline{\includegraphics[width=\columnwidth]{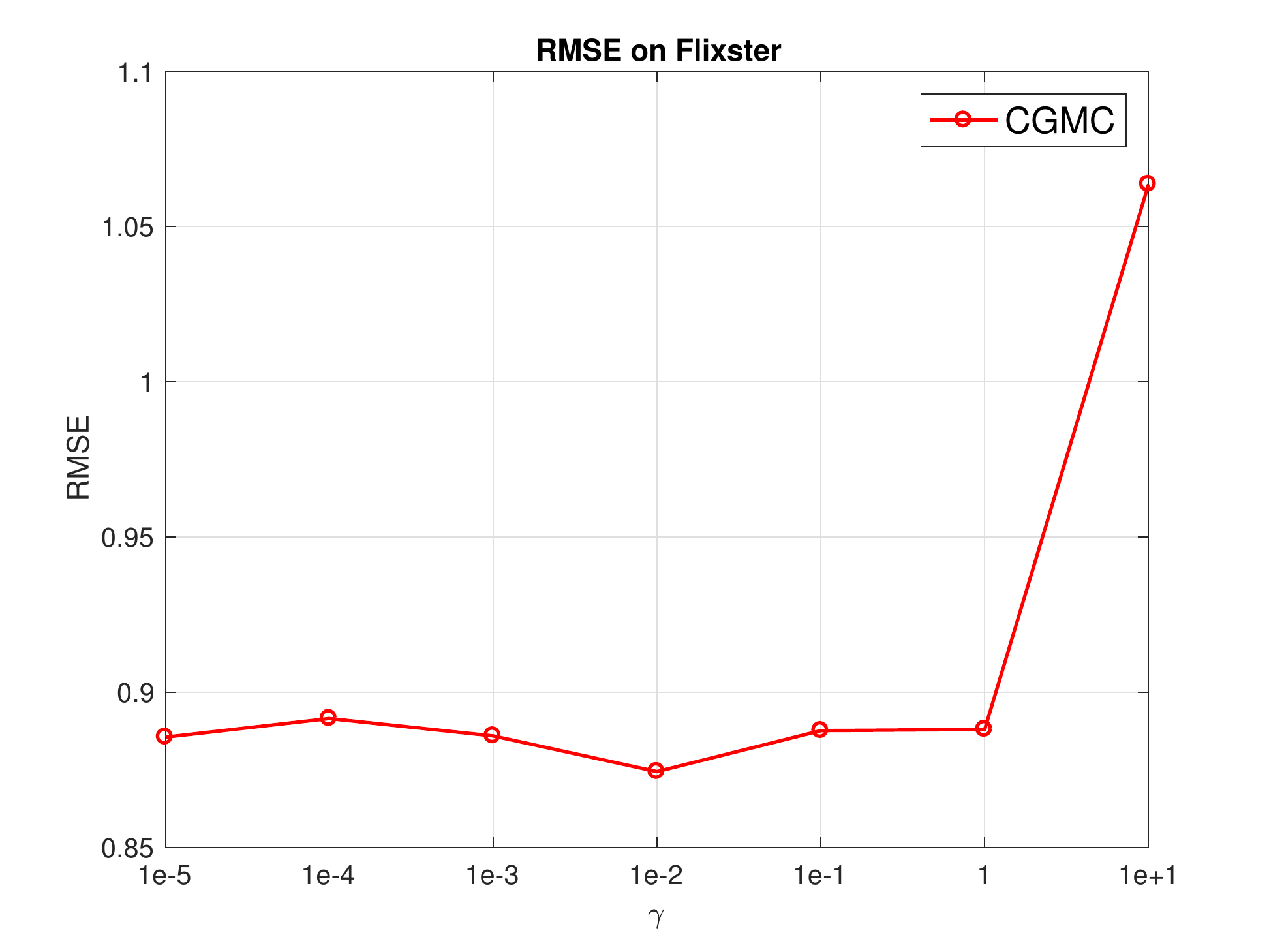}}
			\end{center}
		\end{minipage}
		\hfill
		\begin{minipage}{0.49\linewidth}
			\begin{center}
				\centerline{\includegraphics[width=\columnwidth]{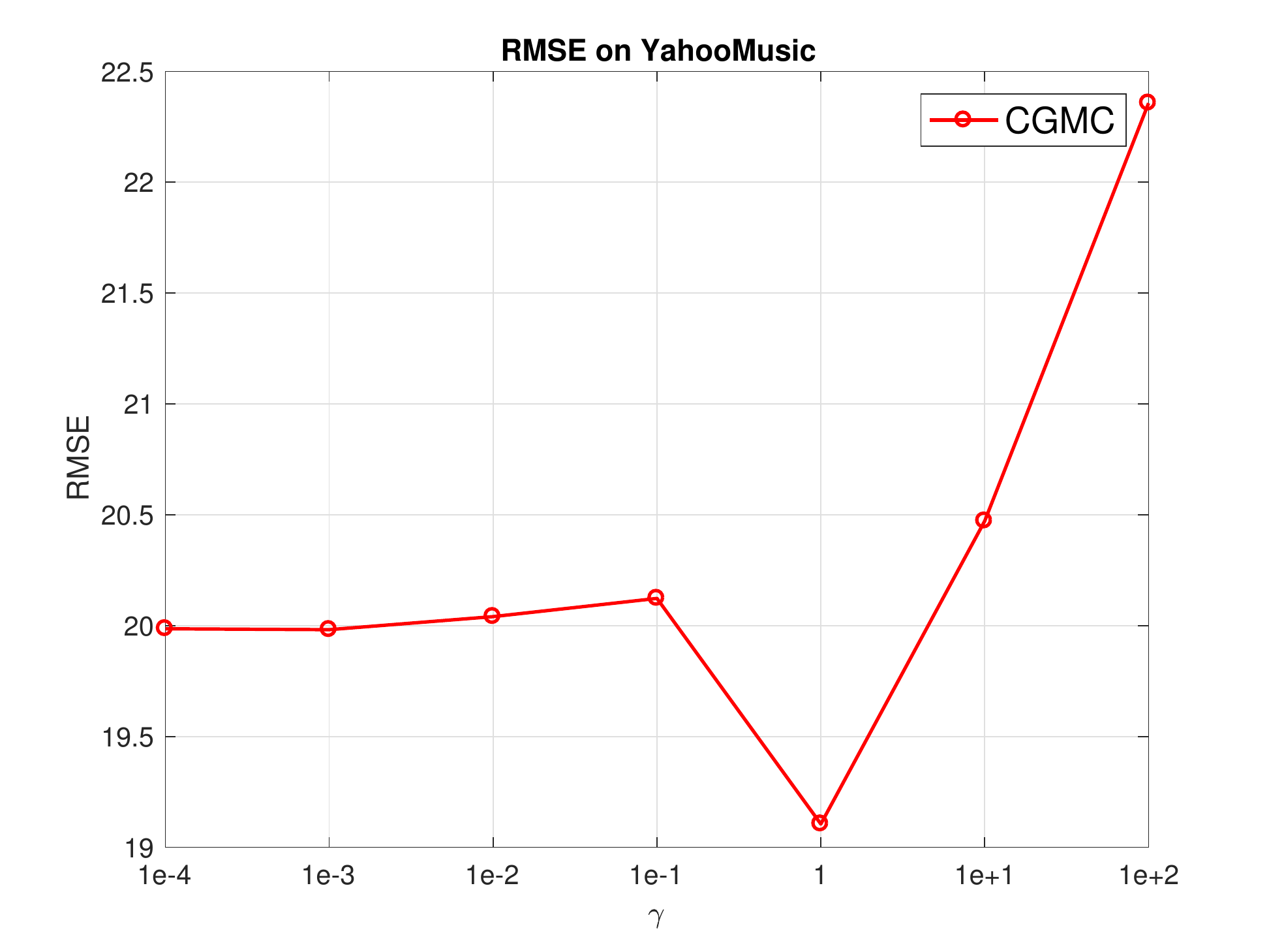}}
			\end{center}
		\end{minipage}
		\vskip -0.2in
		\caption{Sensitivity to $\gamma$.}
		\label{hyper-fig}
	\end{minipage}
	\hfill
	\begin{minipage}{0.5\linewidth}
		\begin{minipage}{0.49\linewidth}
			\begin{center}
				\centerline{\includegraphics[width=\columnwidth]{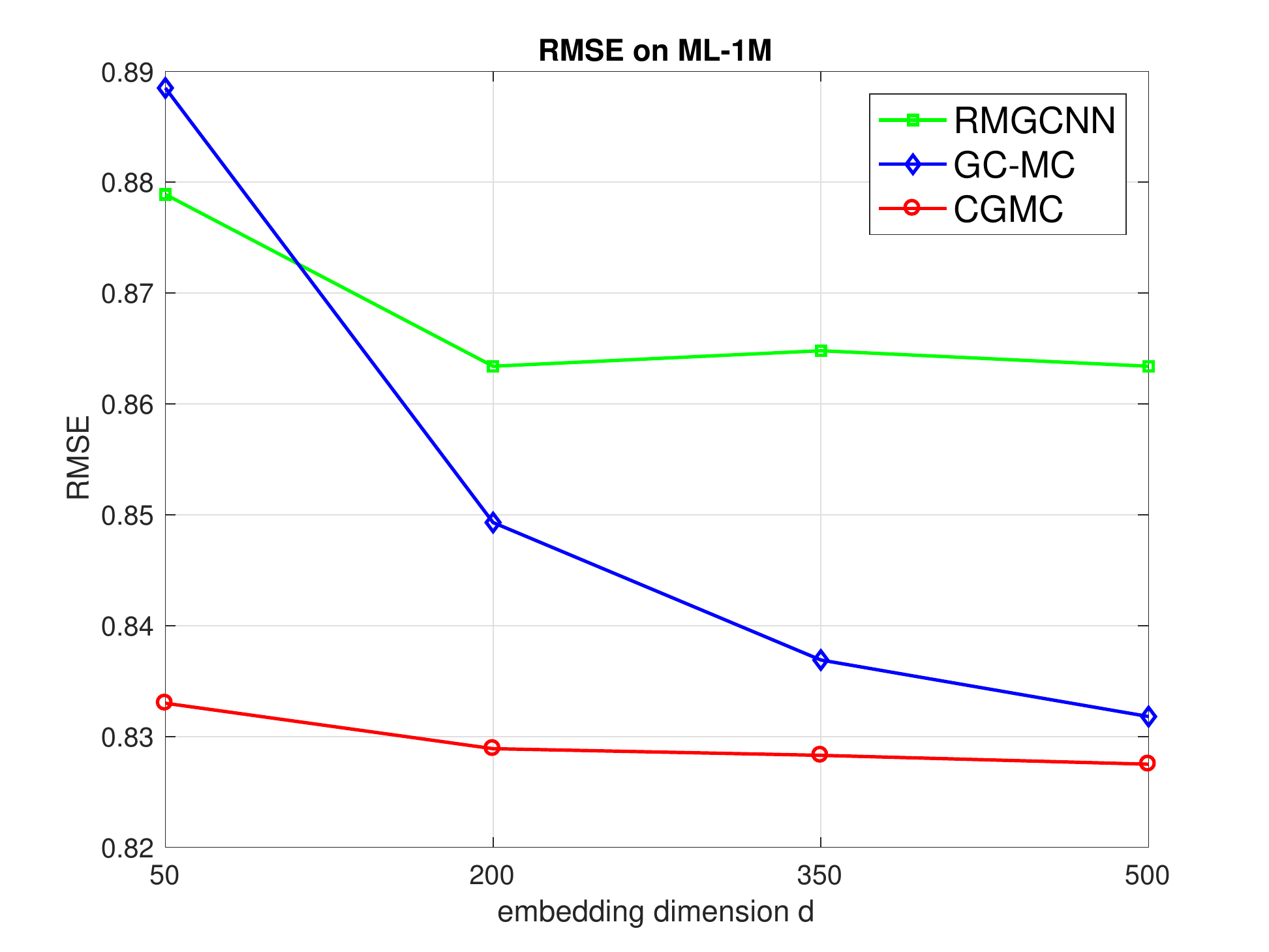}}
			\end{center}
		\end{minipage}
		\hfill
		\begin{minipage}{0.49\linewidth}
			\begin{center}
				\centerline{\includegraphics[width=\columnwidth]{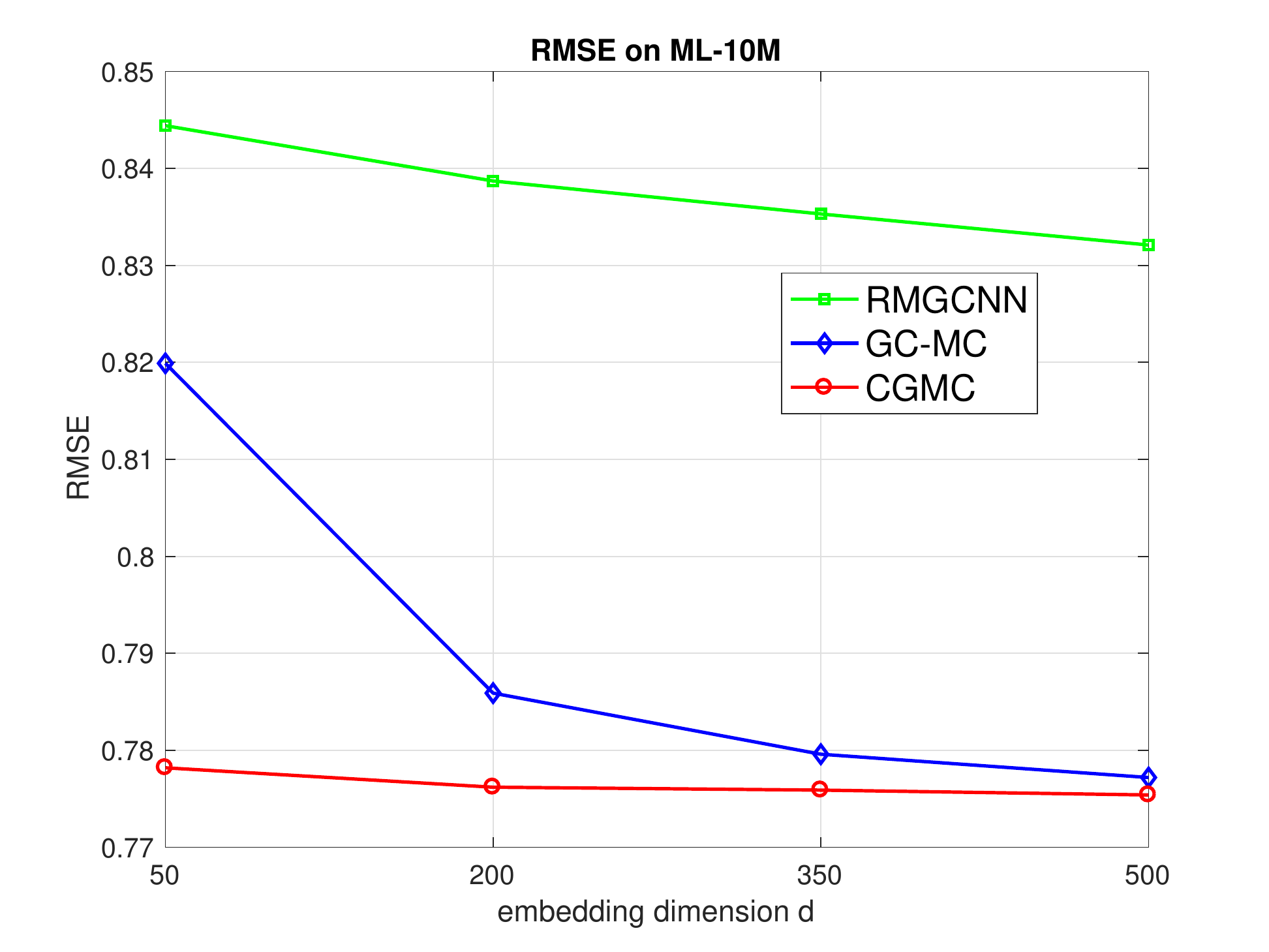}}
			\end{center}
		\end{minipage}
		\vskip -0.2in
		\caption{Effect of embedding dimension $d$.}
		\label{hyper-dim}
	\end{minipage}
\end{figure}

The results are presented in Figure~\ref{hyper-fig} and \ref{hyper-dim}. With respect to $\gamma$, we can see that CGMC behaves well in a wide range of $\gamma$, such as $\gamma \in [10^{-4}, 1.0]$. As for embedding dimension $d$, we train GC-MC with the code provided by the corresponding authors. We can find that CGMC behaves stably on different dimensions and achieves promising results even on low dimension. Moreover, CGMC outperforms the state-of-the-art methods, RMGCNN and GC-MC, on different dimensions.

\section{Conclusion}\label{sec:conclusion}
In this paper, we propose a novel geometric matrix completion (GMC) method, called convolutional geometric matrix completion~(CGMC), for recommender systems with relationship~(link) graphs among users/items. To the best of our knowledge, CGMC is the first work to show that pure graph convolutional network~(GCN) based methods can achieve the state-of-the-art performance for GMC, as long as a proper GCN is designed and a fully connected layer is adopted for space compatibility. Experimental results on four real datasets show that CGMC can outperform other state-of-the-art baselines, including the RMGCNN~\cite{srmgcnn} which is a combination of GCN and RNN.

\medskip

\small
\bibliography{CGMC}
\bibliographystyle{abbrvnat}

\end{document}